\documentclass{new_tlp}
\usepackage{mathptmx}

\usepackage{fancyvrb}

\usepackage{upquote}

\usepackage[breaklinks=true]{hyperref}
\usepackage{breakcites}

\usepackage[justification=centering]{caption}

\newcommand\bcmdtab{\noindent\bgroup\tabcolsep=0pt%
  \begin{tabular}{@{}p{10pc}@{}p{20pc}@{}}}
\newcommand\ecmdtab{\end{tabular}\egroup}

\title[Specifying and Verbalising Answer Set Programs in Controlled Natural Language]
        {Specifying and Verbalising Answer Set Programs in Controlled Natural Language}

  \author[Rolf Schwitter]
         {ROLF SCHWITTER\\
         Department of Computing, Macquarie University, Sydney, NSW 2109, Australia\\
         \email{Rolf.Schwitter@mq.edu.au}}

\jdate{May 2017}
\pubyear{2017}
\pagerange{\pageref{firstpage}--\pageref{lastpage}}
\doi{S1471068401001193}

\begin{document}
\nocite{*}

\label{firstpage}

\maketitle

\begin{abstract}
We show how a bi-directional grammar can be used to specify and verbalise answer set programs 
in controlled natural language. We start from a program specification in controlled natural language 
and translate this specification automatically into an executable answer set program. The resulting 
answer set program can be modified following certain naming conventions and the revised version 
of the program can then be verbalised in the same subset of natural language that was used as 
specification language. The bi-directional grammar is parametrised for processing and generation,
deals with referring expressions,
and exploits symmetries in the data structure of the grammar rules whenever these grammar rules
need to be duplicated. We demonstrate that verbalisation requires sentence planning in order to 
aggregate similar structures with the aim to improve the readability 
of the generated specification. Without modifications, the generated specification is always semantically
equivalent to the original one; our bi-directional grammar is the first one that allows for semantic 
round-tripping in the context of controlled natural language processing. {\em This paper is under 
consideration for acceptance in TPLP}.
\end{abstract}



\section{Introduction}  
There exist a number of controlled natural languages~\cite{Sowa:04,Clark:05,Fuchs:08,Guy:17} that 
have been designed as high-level interface languages to knowledge systems. However, 
none of the underlying grammars of these controlled natural languages is bi-directional in the sense that 
a specification can be written in controlled natural language, translated into a formal target representation, 
and then -- after potential modifications of that target representation -- back again into the same subset of
natural language. This form of round-tripping under modification is difficult to achieve, in particular with a 
single grammar that can be used for processing as well as for generation. In this paper, we show how such a bi-directional 
grammar can be built in form of a logic program for a controlled natural
language. We discuss the implementation of a definite clause grammar that translates a controlled 
natural language specification into an executable answer set program. The resulting answer set program 
can be modified and then verbalised in controlled natural language using the same grammar exploiting
symmetries in the feature structures of the grammar rules. Without modification of the original answer set program,
this leads to a form of round-tripping that preserves semantic equivalence but not syntactic equivalence.
This work has certain similarities with our previous work~\cite{Schwitter:08} where a bi-directional grammar
is used to translate an experimental controlled natural language into syntactically annotated TPTP\footnote{http://www.cs.miami.edu/\~{}tptp/} 
formulas; these annotated formulas are then stored and
serve as templates to answer questions in controlled natural language.
The novelty of the presented work is that no such annotated formulas are required; instead a 
sentence planner is used that applies different generation strategies. The definite clause grammar
uses an internal format for the representation of clauses for the answer set program. This internal
format is designed in such a way that it can be used for processing and generation purposes and
it respects at the same time accessibility constraints of antecedents for referring expressions. The grammar of the controlled 
natural language for answer set programs is more restricted than a corresponding grammar for first-order 
formulas, since some linguistic constructions can only occur at certain positions in particular sentences. However, in
contrast to a controlled natural language that is equivalent to first-order logic, a controlled language
for answer set programs allows us to distinguish, for example, between strong and weak negation 
using two separate linguistic constructions for these forms of negation.

The rest of this paper is structured as follows: In Section 2, we recall linguistic preliminaries and
sketch the controlled natural language to be used. In Section 3, we briefly review answer set
programming and show how it has been employed for natural language processing, with a particular 
focus on controlled natural language processing. In Section 4, we present a specification written in 
controlled natural language, discuss the resulting answer set 
program and its verbalisation. In Section 5, we present the bi-directional 
grammar in detail and explain how symmetries in the feature structures of the grammar rules 
can be exploited to process and generate sentences. In Section 6, we introduce the sentence planner 
and a more complex specification, and discuss the planning strategies that are applied to
the internal format for answer set programs before the bi-directional grammar is used to
verbalise an answer set program. In Section 7, we summarise the key points of our approach, highlight 
its benefits, and conclude.

\section{Linguistic Preliminaries}

The controlled natural language that we discuss here is based on the one used in
the PENG\textsuperscript{ASP} system~\cite{Guy:17}. This system translates 
specifications written in controlled language into answer set programs but not back again.
This is because the grammar of the language PENG\textsuperscript{ASP} has not been designed with bi-directionality 
in mind. The language PENG\textsuperscript{ASP} distinguishes between simple and complex sentences.
We focus in the following introduction mainly on those linguistic constructions that occur also in the subsequent
example specifications.

\subsection{Simple Sentences}

Simple sentences in our controlled natural language consist of a subject and a predicate, and include 
zero or more complements depending on the verbal predicate. For example, the following are simple 
sentences:

\begin{enumerate}
\item Bob is a student.
\item Sue Miller works.
\item Tom studies at Macquarie University.
\item The Biology book is expensive.
\item There is a student.
\end{enumerate}

Sentence (1) consists of a proper name ({\em Bob}) in subject position, followed by a linking verb ({\em is}), 
and a subject complement ({\em a student}). Sentence (2) contains an intransitive verb ({\em works}) but no
complement. Sentence (3) contains a transitive prepositional verb ({\em studies at}) with a complement in
form of a proper name ({\em Macquarie University}) in object position. Sentence (4) consists of a definite 
noun phrase ({\em The Biology book}) in subject position followed by a linking verb and an adjective ({\em expensive}) 
as subject complement. Sentence (5) shows that it is possible to introduce new entities with an expletive 
{\em there}-construction where {\em there} acts as filler for the subject.

\subsection{Complex Sentences}
Complex sentences are built from simpler sentences through coordination, subordination, 
quantification, and negation:

\begin{enumerate}
\setcounter{enumi}{5}
\item Tom studies at Macquarie University and is enrolled in Information Technology.
\item Bob who is enrolled in Linguistics studies or parties.
\item Tom is enrolled in COMP329, COMP330 and COMP332.
\item Every student who works is successful.
\item It is not the case that a student who is enrolled in Information Technology parties.
\item If a student does not provably work then the student does not work.
\end{enumerate}

In sentence (6), two verb phrases are coordinated with the help of a conjunction ({\em and}).
In sentence (7), a subordinate relative clause ({\em who is enrolled in Linguistics}) is used in 
subject position and two verb phrases 
are coordinated with the help of a disjunction ({\em or}). Sentence (8) contains a coordination 
in form of an enumeration ({\em COMP329, COMP330 and COMP332)} in object position. 
Sentence (9) uses a universally quantified noun phrase ({\em Every student}) that contains
a relative clause ({\em who works}) in subject positions. In (10), the entire complex sentence
is negated with the help of the predefined construction {\em It is not the case that}; and finally in (11), weak negation 
({\em does not provably work}) is used in the condition of a conditional sentence and strong 
negation ({\em does not work}) in the consequent of that sentence.

\section{Answer Set Programming and Natural Language Processing}

Answer set programming is a modern approach to declarative programming and has its roots in logic programming,
logic-based knowledge representation and reasoning, deductive databases, and constraint 
solving~\cite{Gelfond:88,Baral:03,Lifschitz:08,Gebser:17}. The main idea behind answer set programming is to
 represent a computational problem as a (non-monotonic) 
logic program whose resulting models (= answer sets) correspond to solutions of the problem. Answer set programs 
consist of clauses that look similar to clauses in Prolog; however, in answer set programming an entirely different computational 
mechanism is used that computes answer sets in a bottom-up fashion. An answer set program consists of a set of clauses (rules)
of the form:

\small
\begin{itemize}
\item[] \texttt{L$_{1}$ ; ... ; L$_{k}$  :- }  \texttt{L$_{k+1}$,  ..., L$_{m}$,}  \texttt{not  L$_{m+1}$,  ..., not  L$_{n}$.}
\end{itemize}
\normalsize

\noindent 
where all \small\texttt{L$_{i}$} \normalsize are literals or strongly negated literals. The connective ``\small\texttt{:-}\normalsize" 
stands for {\em if} and separates the head from the body of the rule. The connective ``\small\texttt{;}\normalsize" in the head 
of the rule stands for an epistemic disjunction (at least one literal is believed to be true). The connective ``\small\texttt{,}\normalsize"
in the body of the rule stands for {\em and} and the connective ``\small\texttt{not}\normalsize" for negation as failure 
(also called weak negation). A rule with an empty body and without an if-connective is called a {\em fact}, and a rule 
without a head is called a {\em constraint}. Answer set programming is supported by powerful tools; for example, the {\em clingo} 
system~\cite{Gebser:17} combines a grounder and a solver. The grounder converts an answer set program that contains 
variables into an equivalent ground (variable-free) propositional program and the solver computes the answer sets 
of the propositional program. Answers to questions can then be extracted directly from the resulting answer sets.

Answer set programming is an interesting formalism for knowledge representation in the context of natural language 
processing because of its ability to combine strong and weak negation which offers support for non-monotonic reasoning. 
Answer set programming has been used for syntactic parsing in the context of the Combinatory Categorial Grammar 
formalism~\cite{Lierler:12,Schueller:13}, for semantic parsing and representation of textual information~\cite{Baral:11,Nguyen:15}, 
and for natural language understanding in the context of question answering~\cite{Baral:06} and the Winograd 
Schema Challenge~\cite{Schueller:14,Bailey:15}. In the context of controlled natural language processing, answer set 
programming has been used as a prototype of a rule system in the Attempto project~\cite{Kuhn:07,Fuchs:08}, as a 
target language for biomedical queries related to drug discovery~\cite{Erdem:09}, as a source language for generating
explanations for biomedical queries~\cite{Erdem:15}, as a framework for human-robot interaction~\cite{Demirel:16},
and as a target language for writing executable specifications~\cite{Guy:17}.

\section{From Controlled Natural Language to Answer Set Programs and Back Again}
Before we discuss the implementation of our bi-directional grammar for answer set programs in Section 5, we first
illustrate how our controlled natural language can be used for specifying and verbalising answer set programs, 
and thereby motivate its usefulness.

\begin{figure}[h!]
\figrule
\begin{verbatim}
  Every student who works is successful.                                    %  C1
  Every student who studies at Macquarie University works or parties.       %  C2
  It is not the case that a student who is enrolled in Information          %  C3    
    Technology parties.
  Tom is a student.                                                         %  C4
  Tom studies at Macquarie University and is enrolled in Information        %  C5
    Technology.       
  Bob is a student.                                                         %  C6
  Bob studies at Macquarie University and does not work.                    %  C7
  Who is successful?                                                        %  C8
\end{verbatim}
\figrule
\caption{{\em Successful Student} in Controlled Natural Language}
\end{figure}

Figure 1 shows a specification written in controlled natural language. This specification consists of two universally 
quantified sentences (\small{\tt C1} \normalsize and \small{\tt C2})\normalsize, a constraint (\small{\tt C3}\normalsize),
four declarative sentences (\small{\tt C4-C7}\normalsize), followed by a {\em wh}-question (\small{\tt C8}\normalsize). 
Note that the two universally quantified sentences contain an embedded relative clause in subject position; the second one 
of these universally quantified sentences (\small{\tt C2}\normalsize) contains a disjunctive verb phrase. The two declarative sentences
(\small{\tt C5} \normalsize and \small{\tt C7}\normalsize) are complex and contain coordinated verb phrases. 
Note also that the proper name \small{\em Macquarie University} \normalsize in \small{\tt C5} \normalsize and \small{\tt C7} \normalsize 
is used anaphorically and links back to the proper name \small{\em Macquarie University} \normalsize  introduced in \small{\tt C2}\normalsize. 
This specification is automatically translated into the answer set program displayed in Figure 2.

\begin{figure}[h!]
\figrule
\begin{verbatim}  
  successful(A) :- student(A), work(A).                                     %  A1
  work(B) ; party(B) :- student(B), study_at(B,macquarie_university).       %  A2
  :- student(C), enrolled_in(C,information_technology), party(C).           %  A3
  student(tom).                                                             %  A4
  study_at(tom,macquarie_university).                                       %  A5                      
  enrolled_in(tom,information_technology).                                  %  A6
  student(bob).                                                             %  A7
  study_at(bob,macquarie_university).                                       %  A8
  -work(bob).                                                               %  A9
  answer(D) :- successful(D).                                               % A10
\end{verbatim}
\figrule
\caption{{\em Successful Student} as Answer Set Program}
\end{figure}

Note that the translation of sentence \small{\tt C1} \normalsize results in a normal clause
(\small{\tt A1}\normalsize) with a single literal in its head. The translation of sentence \small{\tt C2} \normalsize leads to
a clause with a disjunction
 in its head (\small{\tt A2}\normalsize), and the translation of sentence \small{\tt C3} \normalsize to a 
constraint (\small{\tt A3}\normalsize). The translation of the two declarative sentences \small{\tt C4} \normalsize 
and \small{\tt C6} \normalsize results in a single fact (\small{\tt A4+A7}\normalsize) for both sentences, 
and the translation of the two coordinated sentences \small{\tt C5} \normalsize and \small{\tt C7} \normalsize 
leads to two facts each (\small{\tt A5+A6} \normalsize and \small{\tt A8+A9}\normalsize).  In the case 
of \small{\tt C7} \normalsize one of these facts is strongly negated (\small{\tt A9}\normalsize). 
Finally, the translation of the {\em wh}-question \small{\tt C8} \normalsize results in a rule with a specific answer literal
 (\small{\tt answer(D)}\normalsize) in its head (\small{\tt A10}\normalsize). The user can now 
modify this answer set program as illustrated in Figure 3 -- as long as they stick to the existing 
naming convention.

\begin{figure}[h!]
\figrule
\begin{verbatim}  
  stressed(A) :- student(A), work(A).                                      %  A1'
  work(B) ; party(B) :- student(B), study_at(B,macquarie_university).      %  A2'
  :- student(C), enrolled_in(C,information_technology), party(C).          %  A3'
  student(tom).                                                            %  A4'
  study_at(tom,macquarie_university).                                      %  A5'                     
  enrolled_in(tom,information_technology).                                 %  A6'
  student(bob).                                                            %  A7'
  study_at(bob,macquarie_university).                                      %  A8'
  enrolled_in(bob,linguistics).                                            %  A9'
  party(bob).                                                              % A10'
  answer(D) :- stressed(D).                                                % A11'
\end{verbatim}
\figrule
\caption{Modified Answer Set Program}
\end{figure}

This modified answer set program contains the following changes: the literal \small{\tt successful(A)} \normalsize
in the head of the clause \small{\tt A1} \normalsize has been replaced by the literal \small{\tt stressed(A)} \normalsize
on line \small{\tt A1\symbol{13}}\normalsize; the new fact \small{\tt enrolled\_in(bob,linguistics)} \normalsize has 
been added to the program on line \small{\tt A9\symbol{13}}\normalsize; the negated literal \small{\tt -work(bob)} \normalsize 
on line \small{\tt A9} \normalsize has been replaced by the positive literal \small{\tt party(bob)} \normalsize on line
\small{\tt A10\symbol{13}}\normalsize; and the literal \small{\tt successful(D)} \normalsize in the body of the clause 
that answers the question on line \small{\tt A10} \normalsize has been replaced by the literal \small{\tt stressed(D)} \normalsize 
on line \small{\tt A11\symbol{13}}\normalsize. This modified answer set program can now be automatically verbalised in 
controlled natural language as illustrated in Figure 4. 

\begin{figure}[h]
\figrule
\begin{verbatim}
  Every student who works is stressed.                                     %  C1'
  Every student who studies at Macquarie University works or parties.      %  C2'
  It is not the case that a student who is enrolled in Information         %  C3'    
    Technology parties.
  Tom is a student.                                                        %  C4'
  Tom studies at Macquarie University and is enrolled in Information       %  C5' 
    Technology.
  Bob is a student.                                                        %  C6'
  Bob studies at Macquarie University and is enrolled in Linguistics.      %  C7'
  Bob parties.                                                             %  C8'
  Who is stressed?                                                         %  C9'
\end{verbatim}
\figrule
\caption{Verbalisation of the Modified Answer Set Program in Controlled Natural Language}
\end{figure}

We observe that the clauses on line \small{\tt A5\symbol{13}} \normalsize and
\small{\tt A6\symbol{13}} \normalsize and similarly those on line \small{\tt A8\symbol{13}} \normalsize 
and \small{\tt A9\symbol{13}} \normalsize of the modified answer set program have been used to 
generate a coordinated verb phrase as shown on line \small{\tt C5\symbol{13}} \normalsize and
\small{\tt C7\symbol{13}} \normalsize of Figure 4. This is a form of aggregation that has been executed
by the sentence planner and combines similar structures that fall under the same subject based on the proximity of the 
clauses. Alternatively, the sentence planner could use another aggregation 
strategy and combine, for example, the clauses on line \small{\tt A4\symbol{13}} \normalsize and
 \small{\tt A5\symbol{13}} \normalsize or those on line \small{\tt A4\symbol{13}} \normalsize and
 \small{\tt A5\symbol{13}} \normalsize and \small{\tt A6\symbol{13}} \normalsize into a verb phrase 
(more on this in Section 6).

\section{Implementation of the Bi-directional Grammar}
To implement our bi-directional grammar, we use the definite clause grammar (DCG) formalism~\cite{Pereira:80,Pereira:87}. 
This formalism gives us a convenient notation to express feature structures and the resulting grammar can easily 
be transformed into another notation via term expansion, if an alternative parsing strategy is required~\cite{Guy:17}. 
Our bi-directional grammar implements a syntactic-semantic interface where the meaning of a textual specification and its linguistic
constituents is established in a compositional way during the translation of the specification into an answer set program.
Unification is used to map the syntactic constituents of the controlled language on terms and (partial) clauses of the 
corresponding answer set program. In our context, a textual specification $A$ and its verbalisation $A'$ can be shown
to be semantically equivalent, if they both generate the same answer set program that entails the same solutions.

Our implementation of the grammar relies on four other components that are necessary to build an operational system: 
(1) a tokeniser that splits a specification written in controlled natural language into a list of tokens for each sentence; 
(2) a writer that translates the internal format constructed during the parsing process into the final answer set program; 
(3) a reader that reads a (potentially modified) answer set program and converts this program into the internal format; 
and (4) a sentence planner that tries to apply a number of aggregation strategies, reorganises the internal
format for this purpose, before it is used by the grammar for the generation process.

\subsection{Internal Format for Answer Set Programs}

The bi-directional definite clause grammar uses an internal format for answer set programs and a special feature structure 
that distinguishes between a processing and a generation mode in the grammar rules. Depending on this mode,
the same difference list~\cite{Sterling:94} is used as a  data structure to either construct the internal format for an input text
or to deconstruct the internal format to generate a verbalisation. The internal format is based on a flat notation
that uses a small number of typed predicates (e.g. \small{\tt class}\normalsize, \small{\tt pred}\normalsize, \small{\tt prop}\normalsize, 
\small{\tt named}\normalsize) that are associated with linguistic categories (e.g. noun, verb, adjective, proper name) 
in the lexicon. This flat notation allows us to abstract over individual predicate names and makes the processing for the 
anaphora resolution algorithm and sentence planner easier. Let us briefly illustrate the form and structure of this 
internal format with the help of the following two example sentences:

\begin{enumerate}
\setcounter{enumi}{11}
\item Tom is a student and works. 
\item If a student works then the student is successful.
\end{enumerate}

In the case of processing, the tokeniser sends a list of tokens for these two sentences to the bi-directional 
grammar that builds up the following internal format during the parsing process with the help 
of the information that is available in the lexicon for these tokens:

\small
\begin{verbatim}
  [['.', [pred(C,work), class(C,student)], '-:', [prop(C,successful)],
    '.', pred(A,work), class(B,student), pred(A,B,isa), named(A,tom)]]
\end{verbatim}
\normalsize
   
As this example shows, the internal format is built up in reverse order; the literals 
derived from the first sentence (12) follow the literals derived from the second sentence (13). As 
a consequence of this ordering, the arrow operator for the rule (\small{\tt -:}\normalsize) points from the 
body on the left side of the rule to the head on the right side. Note that this internal format 
retains the full stops in order to separate the representation of individual clauses and uses, in our 
example, two sublists as part of the second clause. These sublists constrain the accessibility of 
antecedents for referring expressions -- in a similar way as postulated in discourse representation theory~\cite{Kamp:93,Geurts:15}. 
This internal format is then automatically translated by the writer into the final answer set program. 
In order to achieve this, the writer reverses the order of the clauses, removes
auxiliary literals, and turns the typed representation of literals into an untyped one:

\small
\begin{verbatim}
  student(tom). work(tom).
  successful(C) :- student(C), work(C).
\end{verbatim}
\normalsize

In the case of generation, the clauses of the answer set program are first read by the reader
that reconstructs the flat internal format. Since we use the same data structure (difference list) in 
the grammar for processing and generation, the internal format for clauses is now built up in normal 
order and not in reverse order as before. This has -- among other things -- the effect that the arrow 
operator (\small{\tt :-}\normalsize) of the rule now points from the body of the rule on the right side 
to the head on the left side:

\small
\begin{verbatim}
  [[named(A,tom), pred(A,B,isa), class(B,student), pred(B,work), '.',
    [prop(C,successful)], ':-', [class(C,student), pred(C,work)], '.']]
\end{verbatim}
\normalsize
   
This reconstructed internal format is then sent to the sentence planner that tries to apply a number 
of strategies to aggregate the information in order to generate, for example, coordinated sentences 
or to take care of enumerations in sentences. As we will see in Section 6, sentence planning requires 
reorganising the internal format to achieve the intended result.

\subsection{Overview of Grammar Rules}

In this section we introduce the format of the bi-directional definite clause grammar rules.
 Since our grammar is a grammar for a controlled natural natural, specific restrictions
 apply to the grammar rules. We distinguish between grammar rules for declarative sentences, conditional 
sentences, constraints, and questions. This distinction is important since only certain syntactic constructions 
can occur at certain positions in these sentences. 
For example, a noun phrase in subject position of a declarative 
sentence can only be coordinated if this noun phrase is followed by a linking verb or a noun phrase 
that occurs in the subject position in the consequent of a conditional sentence can only have the form of
a definite noun phrase. 
On the interface level, these syntactic constraints are enforced by a lookahead text editor that 
guides the writing process of the user and uses the grammar as starting point to extract the lookahead information 
(for details about this interface technology see~\cite{Guy:17}). 

When a specification is processed, the text is split up by the tokeniser into a list of lists where each list contains
those tokens that represent a sentence. At the beginning the difference list that constructs the internal representation
for an answer set program has the following form: \small{\tt [[]]-C}\normalsize, where the incoming part consists 
of a list that contains an empty list and the outgoing part consists of a variable; the same applies to the difference 
list that collects all accessible antecedents during the parsing process. When an answer set program is verbalised, 
the incoming part of the difference list consists of an embedded list of elements that stand for the internal format
of the answer set program; of course, the lists for the tokens and accessible antecedents are empty at this point. 
It is important to note that these difference lists are processed recursively for each sentence or 
for each clause of an answer set program and that they establish a context for the actual interpretation.

The following is a grammar rule in definite clause grammar format; it takes as input either a list of tokens of a declarative 
sentence and updates the answer set program under construction or it takes as input a (partial) answer set program and 
produces a list of tokens that correspond to a declarative sentence:

\small
\begin{verbatim}
  s(mode:M, ct:fact, cl:C1-C4, ante:A1-A3) --> 
     np(mode:M, ct:fact, crd:'-', fcn:subj, num:N, arg:X, cl:C1-C2, ante:A1-A2),
     vp(mode:M, ct:fact, crd:'+', num:N, arg:X, cl:C2-C3, ante:A2-A3),
     fs(mode:M, cl:C3-C4).
\end{verbatim}
\normalsize

This grammar rule states that a declarative sentence (\small{\tt s}\normalsize) consists of a noun 
phrase (\small{\tt np}\normalsize) and a verb phrase (\small{\tt vp}\normalsize), followed by a full stop
(\small{\tt fs}\normalsize). This grammar rule contains additional arguments that implement
feature structures in the form of {\em attribute:value} pairs that can be directly processed via 
unification. The feature structure \small{\tt mode:M} \normalsize distinguishes between the 
processing mode (\small{\tt proc}\normalsize) and generation mode (\small{\tt gen}\normalsize).
The feature 
structure \small{\tt ct:fact} \normalsize stands for the clause type and specifies that the noun phrase 
as well as the verb phrase of this rule are used to generate or process facts. The feature structures
 \small{\tt cl:C1-C4}\normalsize, \small{\tt cl:C1-C2}\normalsize, \small{\tt cl:C2-C3}\normalsize, 
and \small{\tt cl:C3-C4} \normalsize implement a difference list and are used to construct the internal 
format for an answer set program or to deconstruct the internal format of an answer set program to 
drive the generation of a verbalisation. The feature structures \small{\tt ante:A1-A3}\normalsize,
\small{\tt ante:A1-A2}\normalsize, and \small{\tt ante:A2-A3} \normalsize implement a second difference list 
that stores all accessible antecedents. These accessible literals are continuously updated by the anaphora
resolution algorithm during the parsing process. The feature structure \small{\tt crd:\symbol{13}-\symbol{13}} \normalsize specifies that 
the noun phrase of this rule cannot be coordinated and the feature structure \small{\tt crd:\symbol{13}+\symbol{13}} \normalsize
 that the verb phrase of this rule can be coordinated. The feature structure \small{\tt num:N} \normalsize deals with number 
agreement between noun phrase and verb phrase, while the feature structure \small{\tt arg:X} \normalsize makes the 
variable for the entity denoted by the noun phrase in subject position (\small{\tt fcn:subj}\normalsize) available 
for the literal described by the verb phrase. Our implementation uses additional feature structures 
that build up a syntax tree and a paraphrase during the parsing process but these feature structures are 
not shown in our example to keep the presentation clear and compact.

The following grammar rule is a preterminal rule that is used in the processing mode (\small{\tt proc}\normalsize) and
adds the literal \small{\tt named(X,Name)} \normalsize for a proper name to the outgoing part of the difference list 
that constructs the internal format and to the difference list that stores all accessible antecedents:

\small
\begin{verbatim}
  pname(mode:proc, fcn:F, num:N, arg:X, 
        cl:[C1|C2]-[[named(X,Name)|C1]|C2],
        ante:[A1|A2]-[[named(X,Name)|A1]|A2]) -->
    { lexicon(cat:pname, mode:proc, wform:WForm, fcn:F, num:N, arg:X, 
              lit:named(X,Name)) },
    WForm.
\end{verbatim}
\normalsize

Note that the outgoing list contains all information of the incoming list plus the added literal. 
At this point, this addition is only preliminary and the anaphora resolution algorithm will decide
if this literal is new or not. The grammar rule that is used in the generation mode
(\small{\tt gen}\normalsize) for proper names looks very similar to the grammar rule in the processing mode, 
but now a literal for the proper name is removed from the incoming list of clauses
 instead of added to the outgoing list:

\small
\begin{verbatim}
  pname(mode:gen, fcn:F, num:N, arg:X, 
        cl:[[named(X,Name)|C1]|C2]-[C1|C2],
        ante:[A1|A2]-[[named(X,Name)|A1]|A2]) -->
    { lexicon(cat:pname, mode:gen, wform:WForm, fcn:F, num:N, arg:X, 
              lit:named(X,Name)) },
    WForm.
\end{verbatim}
\normalsize

The grammar rule below translates conditional sentences into the internal format for rules and 
the internal format for rules into conditional sentences:

\small
\begin{verbatim}
  s(mode:M, cl:C1-C3, ante:A1-A3) -->
     prep(mode:M, wform:['If'], head:H1-H2, body:B1-B2, cl:C1-C2),
     s(mode:M, loc:body, cl:B1-B2, ante:A1-A2),
     adv(mode:M, wform:[then]),
     s(mode:M, loc:head, cl:H1-H2, ante:A2-A3),
     fs(mode:M, cl:C2-C3).
\end{verbatim}
\normalsize

The feature structures \small{\tt cl:C1-C3}, {\tt cl:C1-C2}, and {\tt cl:C2-C3} \normalsize implement again 
a difference list and are used to construct or deconstruct the internal format of a rule. The two feature structures
 \small{\tt cl:H1-H2} \normalsize and \small{\tt cl:B1-B2} \normalsize implement another difference list  
that provides the relevant data structures for the head and body of the rule.
The feature structures \small{\tt ante:A1-A3}\normalsize,
 \small{\tt ante:A1-A2}\normalsize, and \small{\tt ante:A2-A3} \normalsize are used as before to store 
those antecedents that are currently accessible. The overall structure for the rule as well as 
the structure of the head and the body for the rule become available via the following preterminal 
grammar rules:

\small
\begin{verbatim}
  prep(mode:proc, wform:WForm, 
       head:[[]|C2]-[Head, Body, C3|C4], 
       body:[[]|C1]-C2, 
       cl:C1-[[Body, '-:', Head|C3]|C4]) -->
     { lexicon(cat:prep, wform:WForm) },
     WForm.
\end{verbatim}
\begin{verbatim}
  prep(mode:gen, wform:WForm, 
       head:C2-[[]|C1], 
       body:[Body, Head, C3|C4]-[[]|C2], 
       cl:[[Head, ':-', Body|C3]|C4]-C1) -->
     { lexicon(cat:prep, wform:WForm) },
     WForm.
\end{verbatim}
\normalsize

The first rule is used for processing and the second one for generation. In the case of processing a 
conditional sentence, first the body of a rule is constructed and then the head. Constructing the body 
of a rule starts with an empty list (\small{\tt []}\normalsize) and takes the already processed context
in list \small{\tt C1} \normalsize into consideration for anaphora resolution. During the parsing process 
literals are added to the empty list, and the resulting outgoing list (\small{\tt C2}\normalsize) then builds 
the new context for processing the head of the rule. The processing of the head of the rule starts again with an 
empty list and results in an outgoing list that consists of a list (\small{\tt Head}\normalsize) that contains
 the head literals and a list (\small{\tt Body}\normalsize) that contains the body literals. Note that the variable \small{\tt C3} \normalsize 
stands for a list that contains the previously processed context. These elements are then combined in a rule
in the outgoing part of the difference list (\small{\tt [[Body, \symbol{13}-:\symbol{13}, Head|C3]|C4]}\normalsize)
that build up the clauses for the answer set program. It is interesting to see that 
we end up in the case of generation with feature structures that are symmetric to 
processing. In the case of generation, a rule 
is deconstructed into a body (\small{\tt Body}\normalsize) and a head (\small{\tt Head}\normalsize), and then 
the body is first processed followed by the head. Note that the successful processing of the body results in 
an empty list (\small{\tt []}\normalsize) in the outgoing difference list, the same happens after the processing of the head. 
We observe that all grammar rules for preterminal symbols are symmetric, including grammar
rules that process cardinality quantifiers (\small{\tt cqnt}\normalsize) such as {\em exactly Number Noun, at least Number Noun,}  
and {\em at most Number Noun}:

\small
\begin{verbatim}
  cqnt(mode:proc, wform:WForm, num:N, 
       lit:[[]|C]-[Lit|C], 
       cond:[[]|C]-[Cs|C], 
       cl:[[]|C]-[H|C]) -->
     { lexicon(cat:cqnt, mode:proc, wform:WForm, num:N, lit:Lit, cond:Cs, cl:H) },
     WForm.

  cqnt(mode:gen, wform:WForm, num:N,
       lit:[Lit|C]-[[]|C], 
       cond:[Cs|C]-[[]|C],
       cl:[H|C]-[[]|C]) -->
     { lexicon(cat:cqnt, mode:gen, wform:WForm, num:N, lit:Lit, cond:Cs, cl:H) },
     WForm.
\end{verbatim}
\normalsize

The two rules above take part in the processing and generation of clauses with a cardinality constraint in the head:
\small{\tt {\em Lower$_{1}$} {\em \{ Lit:Ds \} Upper$_{2}$ :-} {\em BODY}}\normalsize. Here the expression 
\small{\tt Lit:Ds} \normalsize is a conditional literal consisting of a basic literal (\small{\tt Lit}\normalsize) and one 
or more domain predicates (\small{\tt Ds}\normalsize).

\subsection{Anaphora Resolution}

Proper names and definite noun phrases can be used as referring expressions in the controlled 
natural language PENG\textsuperscript{ASP}. In our case the list structure of the internal representation
constrains the accessibility of antecedents for referring expressions. In particular, a referring expression 
can only refer to an expression in the current list or to an expression in a list superordinate to that list. 
A list that encloses the current list is superordinate to it; and in addition, a list in the body of a rule is
superordinate to the list in the head of the rule, but not vice versa. 

The following grammar rule processes a definite noun phrase and checks whether 
this definite noun phrase introduces a new entity or whether it is used anaphorically. After processing the 
determiner (\small{\tt det}\normalsize) and the noun (\small{\tt noun}\normalsize), the grammar rule calls 
the anaphora resolution algorithm for definite noun phrases  (\small{\tt def}\normalsize) and 
updates the internal format:

\small
\begin{verbatim} 
  np(mode:M, loc:body, crd:'-', fcn:subj, num:N, arg:X, 
     cl:C1-C4, ante:A1-A3) -->
    det(mode:M, num:N, def:'+', cl:C1-C2),
    noun(mode:M, num:N, arg:X, cl:C2-C3, ante:A1-A2),
    { anaphora_resolution(def, M, X, C1, C3, C4, A1, A2, A3) }.
\end{verbatim}
\normalsize

That means after processing of a definite noun phrase and temporarily adding the corresponding
literal (for example, \small{\tt class(X,student)}\normalsize) to the outgoing part (\small{\tt C3}\normalsize) 
of the difference list, the anaphora resolution algorithm checks, whether an accessible literal already 
exists in the data structure \small{\tt C1} \normalsize or not. If one exists, then the variable (\small{\tt X}\normalsize) 
of the newly added literal is unified with the variable of the closest accessible antecedent and 
the temporarily added literal is removed from the outgoing part of the difference list \small{\tt C3}\normalsize; 
resulting in \small{\tt C4}\normalsize. If no accessible antecedent exists, then the definite noun
phrase is treated as an indefinite one and the new literal is not removed. The anaphora 
resolution algorithm also updates the list of accessible antecedents (\small{\tt A3}\normalsize) at 
the same time that can then be used to generate a list of referring expressions for the user interface.

The same grammar rule is used in the case of generation to decide whether a definite noun phrase can 
be generated or not. Since the internal data structure is reduced during the generation process, the anaphora 
resolution algorithm uses the list of accessible antecedents (\small{\tt A1}\normalsize) that is built up in parallel 
to check if a definite noun phrase can be generated or not.

\section{Sentence Planner}

In the case of generation, the reader first reads the answer set program and reconstructs the internal
format. The sentence planner then tries to apply a number of aggregation strategies~\cite{Reiter:00,Horacek:15}. 
Aggregation requires to reorganise the internal format of the answer set program. Remember that the 
internal format relies on a flat notation. In order to reconstruct this internal format for the modified answer 
set program in Figure 3, the reader has to add literals of the form \small{\tt named(Var,{\em Name})} \normalsize 
to the internal representation for each constant (proper name) that occurs in the answer set program. 
After this reconstruction process, the result looks, for example, as follows for \small{\tt A5\symbol{13}} \normalsize and 
\small{\tt A6\symbol{13}}\normalsize:

\small
\begin{verbatim}
  [[..., named(F,tom), 
         pred(F,C,study_at),
         named(C,macquarie_university), '.',
         named(F,tom), 
         prop(F,E,enrolled_in), 
         named(E,information_technology), '.', ...]]
\end{verbatim}
\normalsize

In the next step, the sentence planner tries to apply a suitable aggregation strategy using this
reconstructed representation as a starting point. One possible strategy is to identify identical literals 
that can occur in the subject position (for example, \small{\tt named(F,tom)}\normalsize) and 
literals with the same number of arguments (for example, \small{\tt pred(F,C,study\_at)} \normalsize
and \small{\tt prop(F,E,enrolled\_in)}\normalsize) that can occur in the predicate position, 
and aggregate these literals in order to generate a coordinated verb phrase. In our case, the 
sentence planner generates the following internal structure for verb phrase coordination:

\small
\begin{verbatim}
  [[..., named(F,tom), 
         pred(F,C,study_at), 
         named(C,macquarie_university),
         prop(F,E,enrolled_in), 
         named(E,information_technology), '.', ...]]
\end{verbatim}
\normalsize

The sentence planner only aggregates up to three literals in predicate position and takes
the number of their arguments and their proximity into consideration. That means no
more than three verbs/relational adjectives with the same number of complements are 
currently coordinated in order to achieve good readability of the resulting specification.

In many cases the reconstruction of the internal structure is more complex and the sentence
planner needs to select among the available strategies. We illustrate why this is 
the case with the help of the graph colouring problem~\cite{Gebser:12}.
The graph colouring problem deals with the assignment of different colours to nodes of a graph 
under certain constraints as specified in controlled natural language in Figure 5:

\begin{figure}[h]
\figrule
\begin{verbatim}
  The node 1 is connected to the nodes 2, 3 and 4.                          %  C1  
  The node 2 is connected to the nodes 4, 5 and 6.                          %  C2
  The node 3 is connected to the nodes 1, 4 and 5.                          %  C3
  The node 4 is connected to the nodes 1 and 2.                             %  C4
  The node 5 is connected to the nodes 3, 4 and 6.                          %  C5
  The node 6 is connected to the nodes 2, 3 and 5.                          %  C6
  Red is a colour.                                                          %  C7
  Blue is a colour.                                                         %  C8
  Green is a colour.                                                        %  C9
  Every node is assigned to exactly one colour.                             % C10
  It is not the case that a node X is assigned to a colour                  % C11
    and a node Y is assigned to the colour
    and the node X is connected to the node Y.
\end{verbatim}
\figrule
\caption{{\em Graph Colouring Problem} in Controlled Natural Language}
\end{figure}

This specification shows the use of 
enumerations in object position (\small{\tt C1-C6}\normalsize). It also uses explicit variables
 (\small{\tt X} \normalsize and \small{\tt Y}\normalsize)  in \small{\tt C11} \normalsize to distinguish 
noun phrases that have the same form but denote different entities. The translation of this specification 
results in the answer set program in Figure 6. This program does not keep all the information that is 
available in the internal format. In the internal format, integers ({\small{\tt 1-6}\normalsize}) are represented with the help of 
literals of the form \small{\tt integer({\em Var},{\em Integer})} \normalsize but result in integer
arguments (for example, \small{\tt node(1)}\normalsize) in the answer set program and variables 
 ({\small{\tt X} \normalsize} and {\small{\tt Y}\normalsize}) 
are represented via literals of the form \small{\tt variable({\em Var},{\em VariableName})} \normalsize
which are not necessary anymore in the answer set program after anaphora resolution.

\begin{figure}[t]
\figrule
\begin{verbatim}
  node(1). connected_to(1,2). node(2). connected_to(1,3). node(3).         %  A1'                                       
  connected_to(1,4). node(4).                                          
  connected_to(2,4). connected_to(2,5). node(5). connected_to(2,6).        %  A2'
  node(6).
  connected_to(3,1). connected_to(3,4). connected_to(3,5).                 %  A3'
  connected_to(4,1). connected_to(4,2).                                    %  A4'
  connected_to(5,3). connected_to(5,4). connected_to(5,6).                 %  A5'
  connected_to(6,2). connected_to(6,3). connected_to(6,5).                 %  A6'
  colour(red).                                                             %  A7'
  colour(blue).                                                            %  A8'
  colour(green).                                                           %  A9'
  1 { assigned_to(A,B) : colour(B) } 1 :- node(A).                         % A10'
  :- node(C), assigned_to(C,D), colour(D), node(E), assigned_to(E,D),      % A11'
     connected_to(C,E).
\end{verbatim}
\figrule
\caption{{\em Graph Colouring Problem} as Answer Set Program}
\end{figure}

Generating a verbalisation that starts from the answer set program in Figure 6 requires again first the explicit reconstruction 
of the internal representation by the reader. For example, this reconstruction looks as follows 
for \small{\tt A4\symbol{13}}\normalsize:

\small
\begin{verbatim}
  [[..., class(D,node), integer(D,4), prop(D,A,connected_to), class(A,node), 
         integer(A,1), '.', 
         class(D,node), integer(D,4), prop(D,B,connected_to), class(B,node), 
         integer(B,2), '.', ...]]
\end{verbatim}
\normalsize

The sentence planner then reorganises this internal format in order to prepare for enumerating the 
integers (\small{\tt 1}, \small{\tt 2}\normalsize) of those nodes in object position that have the same subject 
and the same property name. In our case, this results in the following internal representation:

\small
\begin{verbatim}
  [[..., class(D,node), 
         integer(D,4), 
         prop(D,A,connected_to), 
         class(A,node), 
         integer(A,1), 
         prop(D,B,connected_to), 
         class(B,node), 
         integer(B,2), '.', ...]]
\end{verbatim}
\normalsize

The grammar rules are ordered in such a way that priority is given to enumeration for this structure and not
to verb phrase coordination. In our case, the original sentence \small{\tt C4}\normalsize, here
repeated as (a), is generated and not the sentence (b):

\begin{itemize}
\item[a.] The node 4 is connected to the nodes 1 and 2.
\item[b.] The node 4 is connected to the node 1 and is connected to the node 2.
\end{itemize}

In the case of  \small{\tt A11\symbol{13}}\normalsize, the reconstruction process is more complex since
the clause in the answer set program contains two literals ({\small{\tt node(C)} \normalsize and {\small{\tt node(E)}\normalsize}) 
with the same name. These literals need to be distinguished on the level of the controlled natural language with the
help of variable names. That means literals of the form \small{\tt variable({\em Var},{\em VariableName})} \normalsize 
that represent variable names need to be added to the internal representation:

\small
\begin{verbatim}
  [[..., ':-', [class(O,node), variable(O,'X'), 
                prop(O,P,assigned_to), 
                class(P,colour), 
                class(Q,node), variable(Q,'Y'), 
                prop(Q,P,assigned_to), 
                class(P,colour),
                class(O,node), variable(O,'X'),
                prop(O,Q,connected_to), 
                class(Q,node), variable(Q,'Y')], '.', ...]]
\end{verbatim}
\normalsize

In order to generate a verbalisation that distinguishes between indefinite and definite noun phrases in the same way 
as illustrated in \small{\tt C11} \normalsize of Figure 5, the first occurrence of the literals that generate an 
indefinite noun phrase (for example, \small{\tt class(O,node)}\normalsize, \small{\tt variable(O,\symbol{13}X\symbol{13})}\normalsize) 
are added to the outgoing difference list that maintains all accessible antecedents, so that this information
can be looked up later from that list to decide whether a definite noun phrase can be generated or not if 
the same literals occur a second time in the internal representation.

\section{Conclusion}

In this paper, we presented a bi-directional grammar for the controlled natural language PENG\textsuperscript{ASP}.
This grammar uses the definite clause grammar formalism and allows us to translate a specification written in
controlled natural language into an answer set program using a difference list as main data structure. This 
answer set program can then be translated (after potential modifications) back again into a controlled language 
specification. The same grammar rules can be used for processing and for generation; only those grammar rules
that process preterminal symbols and add literals to or remove literals from the difference list need to be duplicated. 
However, these grammar rules have a nice symmetric structure that reflects the processing and generation task. 
The grammar also deals with referring expressions and resolves these expressions during the parsing process. The 
generation process requires sentence planning and allows us to generate semantically equivalent specifications
if the answer set program has not been modified. If the program has been modified, then round-tripping is still
possible as long as we stick to the same naming convention. The presented approach allows domain specialists who
may not be familiar with answer set programming to specify and inspect an answer set program on the level
of a controlled natural language. To the best of our knowledge, this is the first bi-directional 
grammar of a controlled natural language that can be used to process and verbalise answer set programs.

\normalsize

\bibliographystyle{named}

\end{document}